\documentclass[lettersize,journal]{IEEEtran}
\usepackage{amsfonts}
\usepackage{algorithmic}
\usepackage{array}
\usepackage{textcomp}
\usepackage{stfloats}
\usepackage{url}
\usepackage{verbatim}
\usepackage{graphicx}
\usepackage{color}
\usepackage{float}
\usepackage{bm}
\usepackage{amssymb}
\usepackage{threeparttable}
\usepackage{multirow}
\usepackage{eqnarray}
\usepackage{color}
\usepackage{tablefootnote}

\usepackage{amsmath} % assumes amsmath package installed
\usepackage{amssymb}  % assumes amsmath package installed
\usepackage{color}
\usepackage{graphicx}
\usepackage{subfigure}
\usepackage[english]{babel}
\usepackage{amsthm}
\usepackage{breqn}
\usepackage[super]{nth}
\usepackage{url}
\usepackage{siunitx}
\usepackage{makecell}
\usepackage{microtype}
\usepackage{booktabs}
\usepackage{hyperref}
\usepackage{stfloats}
\usepackage{threeparttable} % threeparttable
\usepackage{multirow}
\theoremstyle{definition}

\usepackage{verbatim}

\usepackage[utf8]{inputenc}
\usepackage{graphicx}
\usepackage{float}
\usepackage{bm}
\usepackage{amssymb}
\usepackage{threeparttable}
\usepackage{multirow}
\usepackage{color}
\usepackage{graphicx}
\usepackage{tablefootnote}
\usepackage{algorithm}

\def\BibTeX{{\rm B\kern-.05em{\sc i\kern-.025em b}\kern-.08em
    T\kern-.1667em\lower.7ex\hbox{E}\kern-.125emX}}
\usepackage{balance}
\begin{document}
\title{Non-contact Dexterous Micromanipulation with Multiple Optoelectronic Robots}
\author{Yongyi Jia, Shu Miao, Ao Wang, Caiding Ni, Lin Feng, Xiaowo Wang, and Xiang Li
\thanks{
Y. Jia, S. Miao, X. Wang, and X. Li are with the Department of Automation, Tsinghua University. A. Wang, C. Ni, and L. Feng are with the School of Mechanical Engineering and Automation, Beihang University. This work was supported in part by the Science and Technology Innovation 2030-Key Project under Grant 2021ZD0201404, in part by
Beijing National Research Center for Information Science and Technology, and in part by the National Natural Science Foundation of China under Grant U21A20517 and 52075290. 
Corresponding author: Xiang Li (xiangli@tsinghua.edu.cn)
% Manuscript created October, 2020; This work was developed by the IEEE Publication Technology Department. This work is distributed under the \LaTeX \ Project Public License (LPPL) ( http://www.latex-project.org/ ) version 1.3. A copy of the LPPL, version 1.3, is included in the base \LaTeX \ documentation of all distributions of \LaTeX \ released 2003/12/01 or later. The opinions expressed here are entirely that of the author. No warranty is expressed or implied. User assumes all risk.
}}

% \markboth{Journal of \LaTeX\ Class Files,~Vol.~18, No.~9, September~2020}%
% {How to Use the IEEEtran \LaTeX \ Templates}

\maketitle

\begin{abstract}
 Micromanipulation systems leverage automation and robotic technologies to improve the precision, repeatability, and efficiency of various tasks at the microscale. However, current approaches are typically limited to specific objects or tasks, which necessitates the use of custom tools and specialized grasping methods.
% While much progress has been achieved for robotic manipulation at the macro scale, relatively low dexterity is found in the existing micromanipulation systems, mainly due to the significantly different dynamic models between the macro and micro scales and also the limited sensing and actuation measures. 
% As a result, tasks are with few degrees of freedom (DoFs); 
% microtools have to be customized according to tasks and objects; physical contacts and grasping formulation are also required. 
This paper proposes 
a novel non-contact micromanipulation method based on optoelectronic technologies. The proposed method utilizes repulsive dielectrophoretic forces generated in the optoelectronic field to drive a microrobot, enabling the microrobot to push the target object in a cluttered environment without physical contact. 
% \tcolor{- briefly discuss the noncontact principle, modeling, control, and planning -} 
The non-contact feature can minimize the risks of potential damage, contamination, or adhesion while largely improving the flexibility of manipulation. The feature enables the use of a general tool for indirect object manipulation, eliminating the need for specialized tools.
% the drag effect between the robot and the object, hence reducing the risk of potential damages. 
% The non-prehensile feature can largely improve the flexibility of manipulation, as there is no need to customize specific tools for different tasks. 
A series of simulation studies and real-world experiments---including non-contact trajectory tracking, obstacle avoidance, and reciprocal avoidance between multiple microrobots---are conducted to validate the performance of the proposed method. The proposed formulation provides a general and dexterous solution for a range of objects and tasks at the micro scale.

\end{abstract}


\section{Introduction}
\IEEEPARstart{R}{obot}-assisted micromanipulation has become an emerging field of contemporary scientific research and application \cite{xu2018micromachines}.
%the advancements in microtechnology, biotechnology, and materials science, robotics micromanipulation has garnered considerable attention, holding paramount significance across various disciplines, notably in biomedicine, electronics, and micro/nanofabrication. 
Robotic micromanipulation technology encompasses a spectrum of innovative methodologies, 
% including, but not limited to, external field-driven techniques, micro-pipetting, and microfluidics, 
offering multifaceted solutions for several tasks ranging from targeted drug delivery to intricate microscale assembly, cellular transportation, and single-cell injection \cite{ahmad2021mobile,miao2023microfluidics,dai2022robotic}.
Numerous approaches---including mechanical, fluidic, magnetic, acoustic, electrical, and optical methodologies---have been utilized to automate micromanipulation \cite{sun2022robotic}. However, unlike robotic manipulation at the macro scale, relatively low dexterity has been reflected in the aforementioned micromanipulation systems, as manipulation at the micro scale is characterized by several challenges. For example, the viscous effect is more dominant than inertia, and the sensing and actuation capabilities are limited compared with the capabilities of a robot manipulator. Consequently, the micromanipulation tasks are usually simple and involve few degrees of freedom (DoFs), necessitating the customization of microtools for specific tasks and objects.
\begin{figure}[thb]
    \vspace{-8pt}
    \centering    \includegraphics[width=8.5cm]{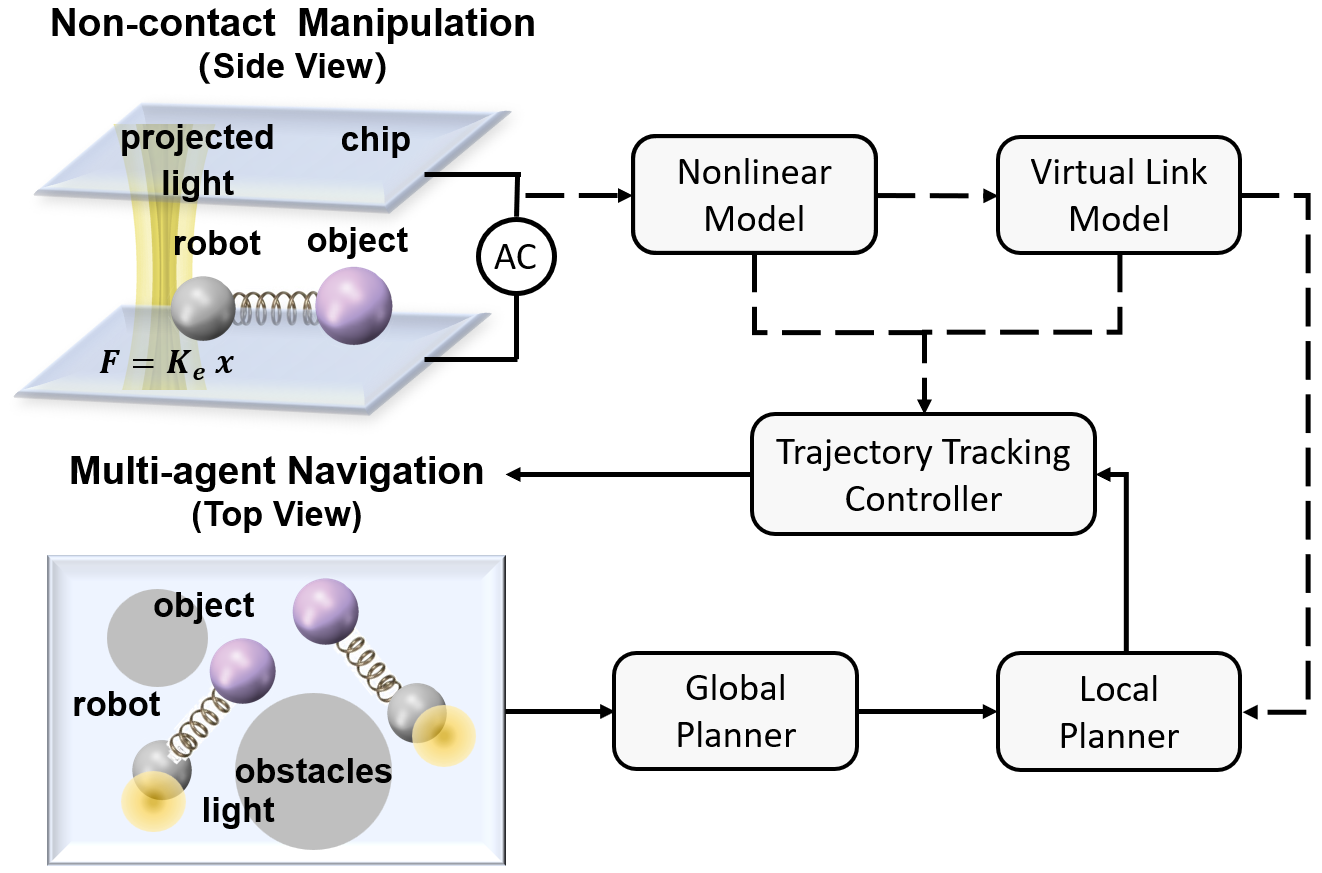}
    \vspace{-5pt}
    \caption{Overview of the proposed framework for non-contact micromanipulation using multiple optoelectronic robots. Projection light-driven general micro-robots use dielectrophoretic repulsion to manipulate targets indirectly. The repulsion can be modeled as a nonlinear model and simplified into a virtual link model. Both global and local planners utilize the virtual link model while considering obstacle avoidance among multiple robot systems and static obstacles. The controller ensures robust tracking of the reference trajectory. (Dashed line: Model loop. Solid line: Closed loop.)}
    \vspace{-12pt}
    \label{fig:fistpage}
\end{figure}

This paper utilizes optoelectronic tweezers (OET) to demonstrate the potential for dexterous manipulation at the micro scale.  
The OET combines optical tweezers with dielectrophoresis and possesses enhanced manipulation force compared with optical tweezers \cite{zhang2022optoelectronic}. Owing to the parallel manipulation capability of the OET, it offers greater potential for applications such as cell sorting and transportation than magnetic control. 
Considering that direct exposure to light in OETs may cause potential phototoxic effects, optoelectronic micro-robots have been designed for indirect manipulation \cite{optoelectronic_microrobot}. These photolithographically manufactured micro-robots are tens of times larger than the targeted objects, thereby diminishing the precision and flexibility of manipulation and potentially leading to mechanical contact with the objects.

Unlike previous studies, this paper introduces a novel approach to the non-contact manipulation of optoelectronic robots, as depicted in Fig. \ref{fig:fistpage}. First, the target objects experience electric dipole repulsion from the robots under an electric field, thereby enabling indirect manipulation without physical contact; this non-contact feature can minimize the risks of potential damage, contamination, or adhesion. Second, 
the microscale robots are arbitrary microspheres capable of being optically trapped without requiring special preparation, and the robots perform the task in a non-prehensile manner, which offers enhanced dexterity. 
Third, 
multiple robots can independently and concurrently transfer target objects to their respective positions with high efficiency and throughput. 

While such a non-contact formulation can provide a general and flexible solution, it also opens up the challenges for modeling, controlling, and planning the micromanipulation system. Specific challenges are presented below.
% The motion control and automatic navigation of this multi-agent underactuated system is not trivial, with its main challenges summarized as follows:
\begin{itemize}
    \item[-] The non-contact and nonlinear dynamics between the robot and the target object, along with unknown parameters and fluid disturbances, present significant challenges for the control and planning of the underactuated system.
    \item[-]  In complex environments with multiple agents, parallel navigation and obstacle avoidance present challenges to the efficiency and feasibility of trajectory planning algorithms.
\end{itemize}

% Note that much progress has been achieved along the direction of non-prehensile manipulation, but the significant gap between different scales hinders the application of those results in the micro world. The non-contact feature is also difficult to achieve with existing solutions. 
To address the above issues and realize dexterous micromanipulation, 
% Numerous recent articles have been proposed on the navigation of microrobots. 
% For instance, extended rapidly-exploring random tree methods and radar-based approaches have been respectively employed for global planning and obstacle avoidance of micromagnetic robot. 
% However, the majority of current research on path planning is concentrated on individual microrobots. In the field of optoelectronic tweezers, some work has been proposed to simultaneously navigate multiple robots, but regardless of dynamic constraints, restricts the applicability of their methods to our task. 
this paper presents a comprehensive planning and control framework for multiple OET-driven robots. The contributions of the paper are summarized as follows.
\begin{itemize}
 \item[-] For motion control, a model predictive control is proposed to indirectly manage the tracking of candidate trajectories for the target object under disturbances. 
 \item[-] For global planning, a multi-agent 
 kinodynamic-RRT* planner guided by curvature is applied to generate collision-free trajectories for the target object. 
 \item[-] For local planning, an optimization-based centralized smoothing method is introduced to provide a locally shortest and reachable trajectory.
\end{itemize} 
A series of simulation studies and real-world experiments---including non-contact trajectory tracking,  obstacle avoidance, and reciprocal avoidance between multiple micro-robots---are conducted to validate the performance of the proposed method. This approach is believed to bridge the gap between progress in the macro world and dexterous micromanipulation. 

\section{Related Works}
% This section reviews the related works on non-contact and non-prehensile micromanipulation.
% \subsection{Non-contact Indirect Micromanipulation}
% Recent studies have explored non-contact manipulation techniques in robotic micro-manipulation to achieve precise control over cells without direct physical contact. These methods utilize various non-contact mechanisms such as acoustic radiation force, magnetic actuation, and dielectrophoresis. For instance, research by Luo et al. (2016) demonstrated non-contact manipulation using a ferrofluid droplet, while Kim et al. (2017) investigated focused acoustic radiation force for single-cell manipulation. Xi et al. (2016) also explored magnetic actuation for non-contact cell manipulation in three-dimensional space. Such approaches offer promising avenues for high-throughput and precise manipulation of cells in microscale environments, as demonstrated by Ma et al. (2016) in ultrahigh-throughput sorting of microfluidic drops. These studies collectively highlight the advancements in non-contact manipulation techniques, enhancing the efficiency and versatility of robotic micro-manipulation in various biomedical and scientific applications.
% \subsection{Optoelectronic Tweezers Manipulation}
OET is an advanced micro-manipulation technique that combines optical tweezers with dielectrophoresis \cite{Massively_parallel}. OET utilizes optical patterns to irradiate photosensitive materials, inducing the generation of non-uniform electric fields in space and thereby polarizing particles to create dielectrophoretic (DEP) forces and enabling the manipulation of micro-particles such as cells, viruses, and large molecules \cite{Trap_profiles, single_DNA}. Compared with optical tweezers, OET systems feature a light-induced dielectrophoretic effect that generates significant manipulation forces at lower light intensities, facilitating parallel manipulation of many small objects by altering projected optical patterns. However, the precision of OET systems is limited by the low imaging precision of optical patterns and the performance of optoelectronic materials, hindering their ability to perform fine manipulations comparable to those of optical tweezers \cite{mi11010078}. 

To avoid optical damage, some indirect manipulation methods involving OET-driven robots have been proposed. For example, micro-gripper structures have been designed to hold and deliver objects \cite{optoelectronic_microrobot}, and electrokinetic adhesion forces between particles have been employed for the transportation and release of objects \cite{Optoelectrokinetics}. However, constrained by fabrication means or physical principles, these indirect manipulation methods cannot accomplish precise and dexterous manipulations or avoid obstacles in narrow scenarios. Direct contact between robots and target objects may also lead to solid contamination, limiting the application of the manipulation methods and further prompting the exploration of non-contact techniques.

Recent studies have explored non-contact techniques in robotic micro-manipulation, aiming to achieve precise control without direct physical contact. The studied methods utilize various non-contact mechanisms, including acoustic radiation force, magnetic actuation, and dielectrophoresis. For instance, Cenev et al. demonstrated non-contact manipulation based on a ferrofluid droplet \cite{cenev2021ferrofluidic}, while Kim et al. investigated a focused acoustic radiation force for single-cell manipulation \cite{kim2021measurements}. Additionally, Icsitman et al. explored magnetic actuation for non-contact magnetic microparticle manipulation in three-dimensional space \cite{icsitman2021non}. These approaches offer promising avenues for high-throughput and precise manipulation of cells in microscale environments, as demonstrated by Lim et al. in the ultrahigh-throughput sorting of microfluidic drops \cite{lim2013ultrahigh}. % These studies collectively highlight the advancements in non-contact manipulation techniques, which enhance the efficiency and versatility of robotic micro-manipulation in various biomedical and scientific applications. 
The existing non-contact studies mostly focus on manipulating individual objects, as in our previous work \cite{10610098}, and the actuation methods and algorithms limit their application to the non-contact manipulation of multiple objects.

% \subsection{Non-contact Indirect Micromanipulation}
%\subsection{Non-prehensile Micromanipulation and Multi-Microrobots Navigation}
%Non-prehensile manipulation \cite{Mason1999ProgressIN} in the macro scale is an important way for robotic manipulation, allowing relative motion between end-effectors and objects instead of tightly grasping \cite{Siciliano2018NonprehensileD}, thus suitable for dexterous manipulation and cluttered scenes. Research endeavors are dedicated to addressing underactuated control and hybrid planning problems in non-prehensile manipulation. Among these, model predictive control \cite{Moura2021NonprehensilePM} and state-triggered constraints \cite{Wang2022ContactImplicitPA} are employed to deal with control in underactuated systems. Monte Carlo tree search\cite{Song2019MultiObjectRW} and reachable-based RRT \cite{10341476} are employed for trajectory generation in handling nonholonomic hybrid dynamics.

% \subsection{Autonomous Navigation for Microrobots}
%Autonomous navigation capabilities in complex and dynamic environments must accomplish practical tasks built upon high-precision closed-loop control. 
In addition, the ability to navigate autonomously in complex environments is a critical step toward the intelligent development of micro-robots. Various macroscopic path planning algorithms have been adapted for microscopic environments, including search-based A* algorithms \cite{fan2018automated}, sample-based rapidly-exploring random trees (RRT) \cite{huang2017path}, and zero-order optimization-based particle swarm optimization (PSO) \cite{wang2021micromanipulation}. Optimality is a key measure of planning algorithms; for instance, the RRT*-connect algorithm can generate collision-free shortest paths for helical magnetic micro-robots \cite{liu20203}. In addition, various local planning algorithms have been proposed to enhance the robustness of planning methods \cite{202100279}, such as combining an A* algorithm with the fuzzy logic method \cite{yang2019automated}. Artificial potential field \cite{lee2021real}, collision-avoidance vector \cite{li2018development}, and radar-based approaches \cite{3263773} have also been employed for local planning. However, most current studies on path planning concentrate on individual micro-robots. Some studies on OETs have explored the simultaneous navigation of multiple robots \cite{9636475, 9841614}. However, these approaches, irrespective of dynamic constraints, are limited in their applicability to non-contact manipulation tasks. 
% \section{System Design}
% \input{chapters/system_design}
% \section{Methodology}
\section{Non-contact Motion Control}
This section proposes a new formulation for non-contact micromanipulation with OET. The approach reduces the risks of physical damage, contamination, and adhesion and offers enhanced dexterity.
\subsection{Non-contact Motion Model}
\begin{figure}[!t]
    \centering    \includegraphics[width=8.5cm]{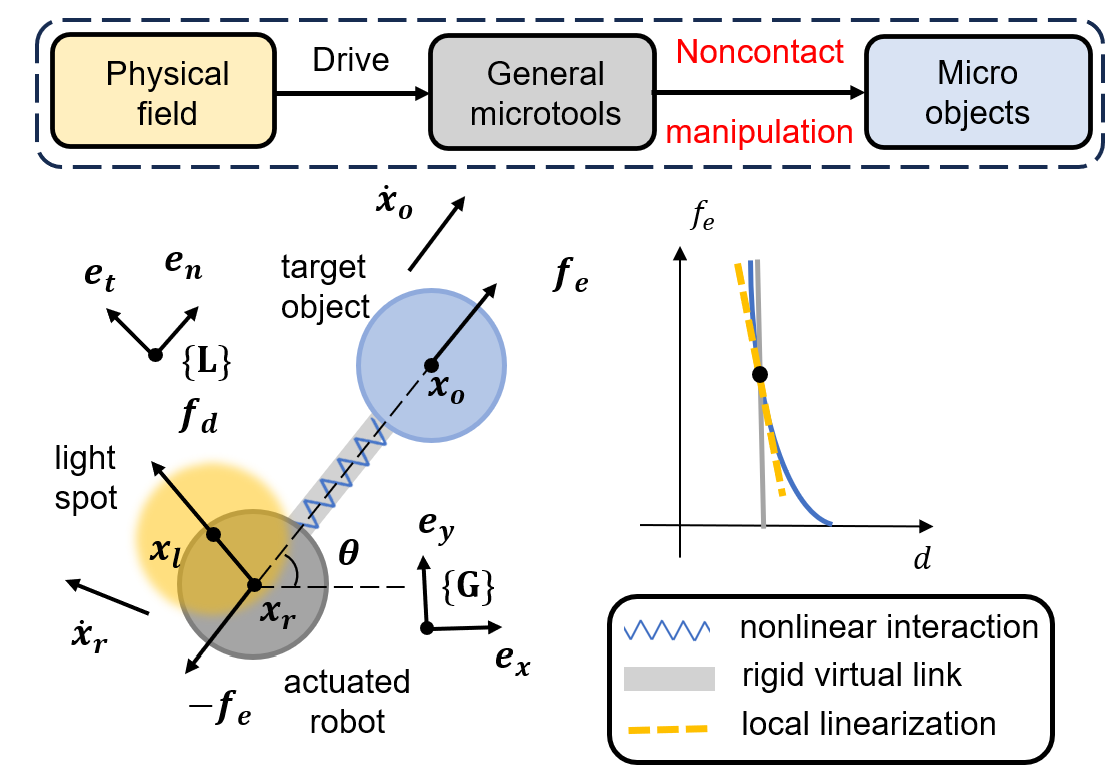}
    \vspace{-8pt}
    \caption{Illustration of an optoelectronic-driven robot performing non-contact manipulation on the target object. The complete nonlinear model, the simplified virtual link model, and the local linear model are proposed and used for control.}
    \vspace{-10pt}
    \label{fig:model}
\end{figure}

In this subsection, we introduce the motion model for non-contact manipulation of a single robot-object pair. As illustrated in Fig. \ref{fig:model}, a pair of a robot and a target object within the OET alternating electric field become polarized, forming induced electric dipoles. A light spot is projected onto the bottom of the OET chip, and a strong electric field region is generated at the center of the photoelectrode. The robot undergoes non-uniform polarization and experiences positive bidirectional dielectrophoretic (p-DEP) forces, which attract the robot toward the high-electric-field gradient. The p-DEP force can be represented as
\begin{equation}
    \bm{f}_{d} = k_d \left(\bm{x}_l - \bm{x}_r\right),
     \label{eq: DEP}
\end{equation}
where $k_d$ is the isotropic stiffness coefficient. Thus, the control input is defined by the displacement between the light spot $\bm{x}_l$ and the actuated robot $\bm{x}_r$, as $\bm{u} = \bm{x}_l - \bm{x}_r \in \mathbb{R}^2$. The homogeneous robot and target object nearby simultaneously experience electrostatic repulsive forces. Because the electric field is perpendicular to the line connecting the centroids of the robot and object, the magnitude of the electrostatic force varies inversely with the fourth power of the distance between them, and the direction of the force aligns along the line connecting the centroids of the robot and object \cite{Electrostatic_Force}. The electrostatic force can be expressed as
\begin{equation}
    \bm{f}_{e} = k_e \frac{\bm{x}_o - \bm{x}_r}{\|\bm{x}_o - \bm{x}_r \|^5},
     \label{eq: Electrostatic_Force}
\end{equation}
where 
%$k_e = \frac{3p^2}{4 \pi \epsilon_o \epsilon_r}$ 
$k_e$ is the electrostatic coefficient. Therefore, the robot driven by the light spot can indirectly control the target object through dielectric force. Neglecting inertia at low Reynolds coefficients, the complete dynamics $\bm{\dot x} = \bm{f}_c \left(\bm{x}, \bm{u}\right)$ is represented by a control-affine model as
\begin{equation}
        \begin{bmatrix}
       \bm{B}_o &      \\
        & \bm{B}_r
    \end{bmatrix}    \begin{bmatrix}
       \bm{\dot x}_o    \\
        \bm{\dot x}_r
    \end{bmatrix} =
    \begin{bmatrix}
       \bm{f}_e (x_o - x_r)     \\
     - \bm{f}_e (x_o - x_r)  
    \end{bmatrix}
    +     
    \begin{bmatrix}
       \bm{0}_2     \\
        k_d\bm I_2
    \end{bmatrix} \bm{u}, 
     \label{eq: whole_nonlinear}
\end{equation}
where $\bm {B}_r, \bm {B}_o \in \mathbb{R}^{2 \times 2}$ are diagonal and positive-definite matrices, representing the damping of the robot and target object; $b_r$ and $b_o$ denote diagonal elements, and $\bm I_2$ is a $2\times 2$ identity matrix. 
The above formulation reflects three characteristics of the system: nonlinearity, self-propulsion, and underactuation, presenting key challenges for tracking control and trajectory generation.

\begin{itemize}
 \item[-] \textbf{Nonlinearity}: The interaction force between the robot and object varies inversely with the fourth power of their distance, introducing nonlinearity into the system. 
 \item[-] \textbf{Self-propulsion}: The repulsive force results in a nonzero autonomous term, necessitating high-frequency controllers to overcome the system's self-propulsion and keep the target object on the desired trajectory.
 \item[-] \textbf{Underactuation}: The robot can only ``push'' the object along the direction of the connecting line with the object rather than ``pull'' it. This requires the controller to consider not only instantaneous actuation but also the long-term consequences of control actions \cite{hogan2020reactive, 10341476}.

\end{itemize} 

Based on the abovementioned characteristics, we propose a simplified model to reduce the system's nonlinearity. Owing to the rapid increase in interaction force as the distance shortens, it can be assumed that a link with infinite stiffness and a length of $d$ exists between the robot and the object (illustrated in the right half of Fig. \ref{fig:model}). We introduce a local reference frame fixed at the center of the robot, with its normal direction aligned along the link. In the local frame, the normal velocity $v_t$ of both robot and object is the same, while the object has no tangential velocity $v_n$. Consequently, the simplified kinematics is expressed as
\begin{equation}
    \bm{\dot x}_s =  
    \begin{bmatrix}
       {\dot x}_{o,x}    \\
       {\dot x}_{o,y}     \\
        \dot \theta
    \end{bmatrix} =
    \bm{J} 
    \begin{bmatrix}
       v_n    \\
       v_t
    \end{bmatrix} , 
     \label{eq: kinematics}
\end{equation}
where $\theta$ is local/global orientation angle, and $\bm{J}$ denotes the  kinematic Jacobian matrix, defined as 
\begin{equation}
    \bm{J} =
    \begin{bmatrix}
       \cos \theta & 0  \\
       \sin \theta & 0  \\
       0 & - \frac{1}{d} 
    \end{bmatrix} .
     \label{eq: Jacobian}
\end{equation}
The dynamics of the two base directions can be decoupled as $v_n = \frac{k_d}{b_o+b_r} u_n $ and $v_t = \frac{k_d}{b_r} u_t$. From the above equations, the simplified dynamic is described as
\begin{equation}
    \bm{\dot x}_s = \bm{J}\bm{K}_d\bm{R} \bm{u},
    \label{eq: simplified_dynamic}
\end{equation}
with the stiffness matrix $\bm{K}_d = \text{diag}(\frac{k_d}{b_o+b_r}, \frac{k_d}{b_r}) \in \mathbb{R}^{2 \times 2}$ and rotation matrix $\bm{R} \in \mathbb{SO}^2$.
Compared with (\ref{eq: whole_nonlinear}), the simplified model 1) eliminates the autonomous term and 2) reduces the state dimensionality and nonlinearity. The position of robots can be obtained as
\begin{equation}
    \bm{x}_r = \bm{G} \bm{x}_s =     \begin{bmatrix}
       1 & 0 &  -d \cos{\theta} \\
       0 & 1 &  -d \sin{\theta}
    \end{bmatrix} \bm{x}_s.
    \label{eq: position_robot}
\end{equation}

\subsection{Trajectory Tracking Controller} \label{sec.ttc}
\begin{figure}[!h]
    \centering    \includegraphics[width=8.5cm]{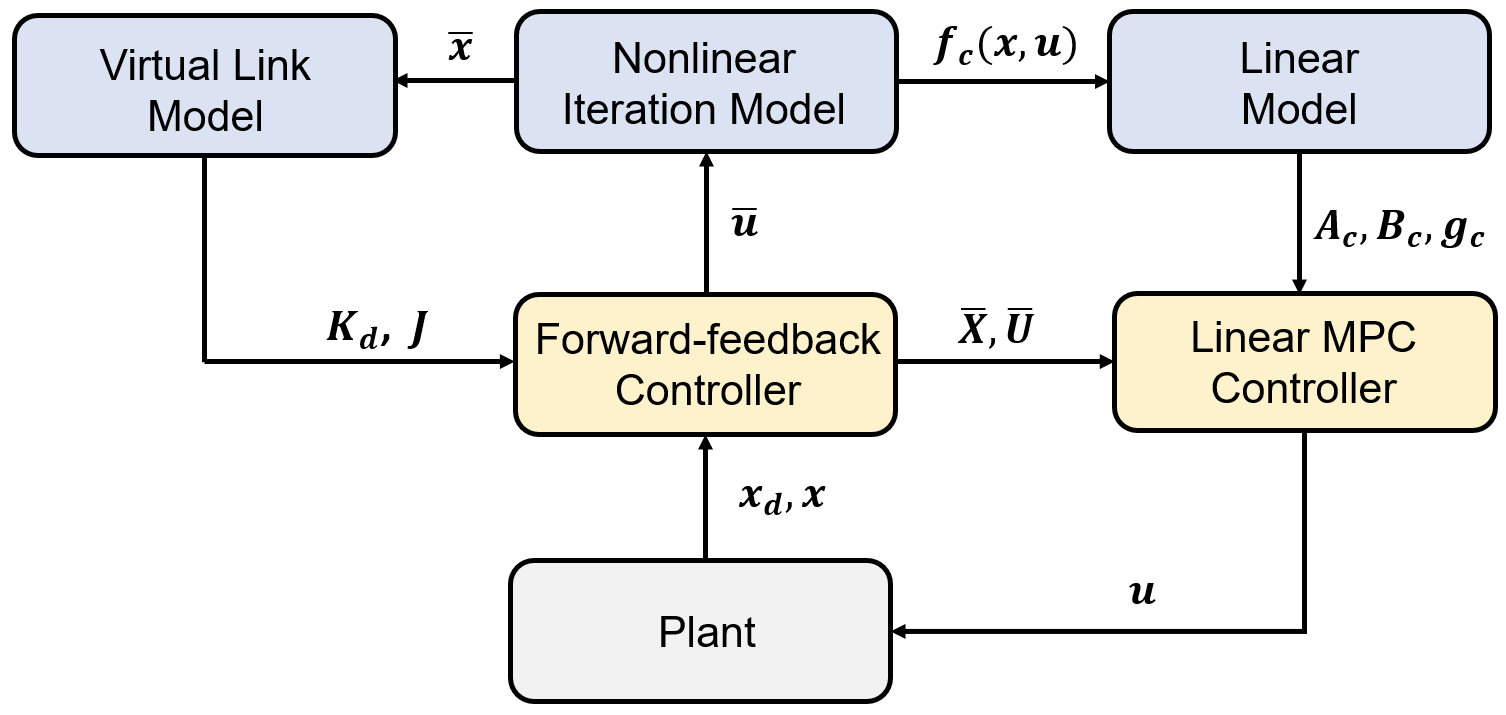}
    \vspace{-8pt}
    \caption{Control diagram of the proposed method. The FF--FB controller generates the nominal trajectory, and the linear MPC controller performs fine-tuning.}
    \vspace{-10pt}
    \label{fig:control}
\end{figure}

In this subsection, we present a method of non-contact controlling a single target object to track a target reference trajectory denoted as $\bm{x}_o^{\text{ref}}$. 
%The target reference trajectory denoted as $\bm{x}_o^{\text{ref}}$ could be specified either manually or by the trajectory generator introduced in the next section. 
Given the underactuated and self-propelled characteristics of the systems delineated in the preceding section, a controller is needed to harmonize forward prediction with computational speed. Owing to the impracticable computational demands of directly solving nonlinear MPC, as described in Fig. \ref{fig:control}, we initially deploy a feedback controller to generate a nominal trajectory. Subsequently, we utilize the linearized model along the trajectory to compute model predictive control (MPC) inputs. 

First, a feedforward--feedback (FF--FB) controller is designed as follows according to the simplified model (\ref{eq: simplified_dynamic}) \cite{10.5555/3165183}
\begin{equation}
    \bm{\overline{u}} = \left(\bm{K}_d\bm{R}\right)^{-1} \bm{J}^{\dag} (\bm{K}_p \Delta \bm{x}_s +  \bm{\dot x}_s^{\text{ref}}),
    \label{eq: p_controller}
\end{equation}
where {the simplified reference} $\bm{x}_s^{\text{ref}}$ is denoted as $\left[\bm{x}_o^{\text{ref}}; \arctan \left({\bm{x}_{o,y}^{\text{ref}}},{\bm{x}_{o,x}^{\text{ref}}}\right)\right]$, the tracking error $\Delta \bm{x}_s$ is calculated as $\bm{x}_s^{\text{ref}} - \bm{x}_s$, and the $\bm{K}_p \in \mathbb{R}^3$ is the proportional gain matrix. $\bm{J}^{\top}\bm{J}$ is an identity matrix from (\ref{eq: Jacobian}), so the pseudo-inverse $\bm{J}^{\dag}$ can be simplified to $\bm{J}^{\top}$. 

The state of next time step $\bm{\overline x}_{k+1}$ can be obtained through forward Runge--Kutta simulation using the control input and the complete model (\ref{eq: whole_nonlinear}), as
\begin{equation}
    \bm{\overline x}_{k+1} = \bm{f}_c^{\text{RK4}} (\bm{\overline{x}}_{k}, \bm{\overline{u}}_{k}, \delta t).
    \label{eq: RK4}
\end{equation}
Through iterative utilization of (\ref{eq: p_controller}) and (\ref{eq: RK4}), a nominal trajectory of horizon length $H$ can be derived as $\bm{\overline{x}}_0, \bm{\overline{u}}_0, \cdots, \bm{\overline{x}}_{H-1}, \bm{\overline{u}}_{H-1}, \bm{\overline{x}}_{H}$, where $\bm{\overline{x}}_0$ is equal to the current state $\bm{x}_0$ . The linearization of the complete model (\ref{eq: whole_nonlinear}) about the nominal trajectory is computed as
\begin{equation}
    \bm{\dot x} = \bm{A}_c \bm{\delta x} + \bm{B}_c \bm{\delta u} + \bm{g}_c,
    \label{eq: linear}
\end{equation}
where $\bm{\delta x} = \bm{x} - \bm{\overline x}$, $\bm{\delta u} = \bm{u} - \bm{\overline u}$, and
$
    \bm{A}_c=\frac{\partial}{\partial {\bm{x}}}\bm{f}_c({\bm{x}},{\bm{u}})\Big|_{(\bar{\bm{x}},\bar{\bm{u}})}, \bm{B}_c=\frac{\partial}{\partial \bm{u}}\bm{f}_c(\bm{x},\bm{u})\Big|_{(\bar{\bm{x}},\bar{\bm{u}})}, \bm{g}_c = \bm{f}_c(\bar{\bm{x}},\bar{\bm{u}}).
    \label{eq: rk4}
$
In the presence of initial biases and disturbances, the linearized model obtained through this approach is closer to the actual model than that linearized directly from the reference trajectory. Therefore, the optimal control input can be defined as the solution to the following quadratic optimization problem:
\begin{equation}
\begin{aligned}
     \min\limits_{\mathcal{U}=\{\bm{u}_0, \cdots , \bm{u}_{H-1}\}} & \mathcal{J} = \sum_{k=0}^{H} \bm \|\bm{x}_{o,k} - \bm {x}_{o,k}^{\text{ref}}\|_{\bm{Q}}  + \|\delta \bm{u}_k \| _{\bm R} \\
    s.t. \quad \bm{x}_{k+1} = & \bm{x}_k + \delta t \left[\bm{A}_{c,k} \bm {\delta x}_k + \bm{B}_{c,k} \bm {\delta u}_k + \bm{g}_{c,k}\right], \\
    \bm{\delta x}_k =& \bm{x}_k - \bm{\overline x}_k , \bm{\delta u}_k = \bm{u}_k - \bm{\overline u}_k, \bm{\delta x}_0 = \bm{0},  \\ 
     \quad \left| \bm {u}_{k} \right| \leq & \bm {u}_{\text{max}}, \quad k = 0, \cdots, H-1\\
\end{aligned}
    \label{eq: optimal}
\end{equation}
where the $\bm{Q}, \bm{R}$ are weighting matrices, and the related terms are
designed to penalize the deviation of the target object from the reference trajectory and the magnitude of the control input. 
Owing to the localized effectiveness of the dielectrophoretic force, we constrain the maximum value of the input. The multi-shooting approach is utilized to construct dynamic constraints, and the first control input $\bm{u}_0 $ is extracted and sent to the OET projection system.

\section{Multi-agent Path Planning}
The capability for parallel manipulation is a significant advantage of OETs compared with other micromanipulation methods \cite{10611608}, and it allows for a multi-agent environment consisting of multiple robot--object pairs. This paper proposes a coarse-to-fine efficient planner to generate reference trajectories for each target object, enabling multi-agent navigation in clutter scenarios. 

\subsection{Global Path Planning}
For global path planning, we propose a multi-agent curvature-optimized rapidly exploring random tree star (MACO-RRT*) algorithm that avoids collisions between objects and environmental obstacles while considering dynamics reachability, given the initial states $\bm{x}_{\rm init}$ and target positions $\bm{x}_{o,\rm targ}$. 

As described in Algorithm 1, an exploring tree $T_n$ is maintained for the $n^{th}$ robot--object pair. At each round of extension, a random point $\bm{x}_{\rm{rand}}$ is first sampled in two-dimensional space $\bm{X} \in \mathbb{R}^2$. To consider the non-holonomic kinematics in (\ref{eq: kinematics}), the local/global orientation angle $\theta$ is approximately introduced as 
\begin{equation}
\theta  = \arccos \left(\frac{\bm{x}_{{\rm cur}}\cdot \bm{x}_{{\rm par}}}{\|\bm{x}_{{\rm cur}}\|_2 \|\bm{x}_{{\rm par}}\|_2}\right),
\label{eq: turn angle}
\end{equation}
where $\bm{x}_{\rm{cur}}$ is the position of the current node, and $\bm{x}_{\rm{par}}$ is the position of the parent node. The distance metric in the $FindNearest$ function can be expressed as
\begin{equation}
Dist(\bm{x}_{\rm{rand}}, \bm{x}) = \left |\arccos\left(\frac{\text{tr}\left(\bm{R}_{\rm {rand}}^\intercal\bm{R}\right)}{2}\right) \right| + \lambda\|\bm{x}_{\rm {rand}} - \bm{x}\|_2,
\label{eq: distance}
\end{equation}
where $\bm{R}$ represents the rotation matrices of orientation angles and $\lambda$ is a proportional coefficient. A new state $\bm{x}_{\rm{new}}$ is chosen by the $Steer$ function with a small step. In addition to considering collisions with the environment, robot--object interactions are also detected, manifested in the form of repulsive forces as (\ref{eq: Electrostatic_Force}). Owing to the decay characteristics of the repulsive force, we assume a radius of the effective range around a robot or an object, denoted as $r_e$, and then define ``collision-free'' as outside of this range. The $InterCollisionFree$ function uses the path index from a node to the root to check for collisions with nodes having nearby indices in other trees. After $\bm{x}_{\rm{new}}$ is inserted into the tree by function $InsertNode$, the surrounding nodes $\bm{x}_{\rm{neig}} \in \bm{X}_{\rm{neig}}$ are evaluated to determine if they can be reached with a lower cost via a motion involving $\bm{x}_{\rm{new}}$ without collision. The node cost is calculated using (\ref{eq: distance}) as 
\begin{equation}
Cost(\bm{x}_{\rm{cur}}) = Cost(\bm{x}_{\rm{par}}) + Dist(\bm{x}_{\rm{cur}}, \bm{x}_{\rm{par}}).
\label{eq: cost}
\end{equation}
Nodes that meet the above conditions are rewired by the function $RewireConnect$. When $\bm{x}_{\rm {new}}$ is close enough to the target position, the expansion of the current tree terminates. After all trees stop expanding, initial feasible paths are found for all objects.

\begin{algorithm}
\caption{MACO-RRT* }
\label{alg:rrt}
\begin{algorithmic}[1]
    \STATE \textbf{Input:} number of robot-object pairs $N$, initial states $\bm{x}_{\rm{init}}$, target positions $\bm{x}_{o, \rm{targ}}$, sampling space $\bm{X}$, maximum tree size $N_{\rm{max}}$
    \STATE \textbf{Output:} collision-free paths from $\bm{x}_{\rm{init}}$ to  $\bm{x}_{o, \rm{targ}}$
    \FOR{$n = 1$ \textbf{to} $N$}
    \STATE $T_n \leftarrow InitTree(\bm{x}_{\rm{init}, n})$;
    \ENDFOR
    \FOR{$i = 1$ \textbf{to} $N_{\rm{max}}$}
        \FOR{$n = 1$ \textbf{to} $N$}
        \STATE $\bm{x}_{\rm{rand}} \leftarrow RandomNode(\bm{X})$;
        \STATE $\bm{x}_{\rm{near}} \leftarrow FindNearest(\bm{x}_{\rm{rand}}, T_n)$; 
        \STATE $\bm{x}_{\rm{new}} \leftarrow Steer(\bm{x}_{\rm{near}}, \bm{x}_{\rm{rand}})$;
        \IF{$EnvCollisionFree(\bm{x}_{\rm{new}})$ and \\ \quad $InterCollisionFree(\bm{x}_{\rm{new}}, T, i)$} 
        \STATE $T_n \leftarrow InsertNode(\bm{x}_{\rm{near}}, \bm{x}_{\rm{new}}, T_n, i);$
        \STATE $T_n \leftarrow Rewire(\bm{x}_{\rm{new}}, T, i);$
        \ENDIF
        \ENDFOR
    \IF{$FindPath(T)$}
    \RETURN $T, path$
    \ENDIF
    \ENDFOR
    
\STATE \textbf{function} $Rewire(\bm{x}_{\rm{new}}, T_n, i)$ 
\STATE $\quad \bm{X}_{\rm{neig}} = FindNeighbors(\bm{x}_{\rm{new}}, T_n);$
\STATE \quad \textbf{for} {$\bm{x}_{\rm{neig}} \in \bm{X}_{\rm{neig}}$}
\STATE \quad \quad \textbf{if}  $InterCollisionFree(\bm{x}_{\rm{neig}}, T, i+1)$ \textbf{and} \\ 
\quad \quad $Cost(\bm{x}_{\rm{new}}) + Dist(\bm{x}_{\rm{new}}, \bm{x}_{\rm{neig}}) \leq Cost(\bm{x}_{\rm{neig}})$
\STATE \hspace*{2.25em} $T_n = RewireConnect(\bm{x}_{\rm{new}}, \bm{x}_{\rm{neig}})$
\STATE \hspace*{2.25em} $Cost(\bm{x}_{\rm{neig}}) = Cost(\bm{x}_{\rm{new}}) + Dist(\bm{x}_{\rm{new}}, \bm{x}_{\rm{neig}})$
\STATE \textbf{return}
\end{algorithmic}
\end{algorithm}
\vspace{-15pt}
\subsection{Local Trajectory Generation}
% The capability for parallel manipulation constitutes a significant advantage of optoelectronic tweezers over other micromanipulation methods \cite{}. This characteristic results in a multi-agent environment consisting of numerous robot-object pairs. Therefore, a trajectory planner is required to generate reference trajectories for each object to avoid mutual interaction between different objects and robots. The interaction is also manifested in the form of repulsive forces as (\ref{eq: Electrostatic_Force}). Due to the decay characteristics of the repulsive force, we assume a radius of influence around a robot or object, denoted as $r_c$, and then define the “collision-free” out of this range. 

%Furthermore, the consideration of dynamic is essential as underactuated systems cannot track arbitrary trajectories. Given these requirements, an offline planner is proposed to generate a reference trajectory $\bm x_o^{\text{ref}}$ that simultaneously satisfies both the dynamic and collision avoidance constraints, given the initial state $\bm {x}^{\text{init}}$ and target objects' position $\bm x^{\text{target}}$.
The coarse path obtained by the above sampling-based method is discrete and needs to be smoothed to improve trackability. Hence, an optimization-based smoothing method is introduced to provide a locally shortest trajectory.

\textit{1) Collision-Free Constraints}: 
First, obstacle avoidance for two agents is considered within a time step $\left[t, t+\delta t\right]$. To address the high computational burden associated with a fine discretization of the trajectory, a polyhedral outer representation is utilized to approximate a segment of the trajectory \cite{tordesillas2019faster, zhou2019robust}. The outer polyhedral is typically represented by the convex hull of control points that parameterize the trajectory. 

\begin{figure}[tb]
    \centering
    \includegraphics[width=8.5cm]{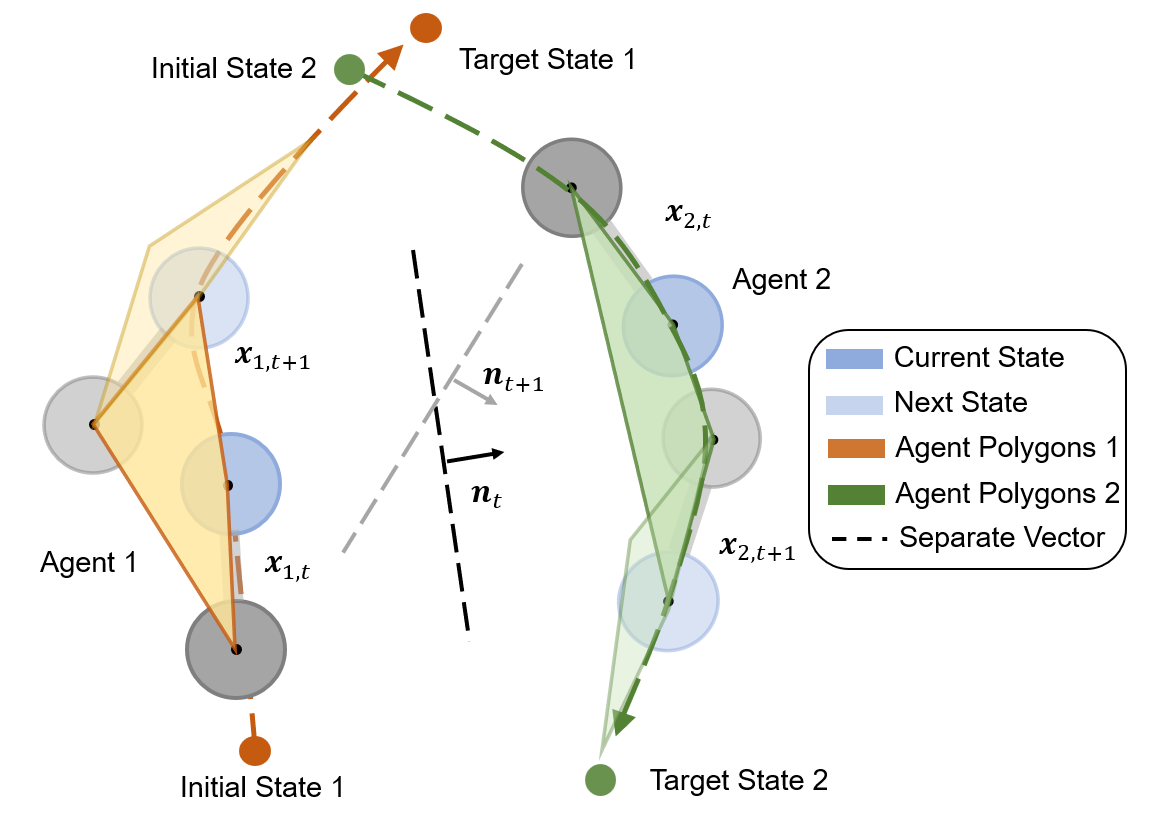}
    \vspace{-8pt}
    \caption{Illustration of the collision-free constraints between different agents. In each time period, the trajectories of the two robot--object pairs are represented as polyhedral and separated by a vector.}
    \vspace{-12pt}
    \label{fig:constraints}
\end{figure}

In this paper, we generate convex outer polygons using the positions of the robot $\bm{x}_r$ and target object $\bm{x}_o$ at two adjacent time steps (a total of four control points whose set is denoted as $\mathcal{P}$), as illustrated in Fig. \ref{fig:constraints}. The robot and the object during this time interval can be approximated to be located within the convex polygon $\mathcal{C}$. The convex polyhedrons of the $i \text{th}$ and $j \text{th}$ agents can be separated by a vector represented by the normal $\bm{n}_{ij}$ and distance from the origin $d_{ij}$. Therefore, the collision avoidance condition is formulated as follows:

\begin{equation}
    \begin{aligned}
    &\bm{n}_{ij}^{\top}\bm{p}_i + d_{ij} > \frac{r_e}{2} \quad \forall \bm{p}_i \in {\mathcal{P}_i},  \\
    &\bm{n}_{ij}^{\top}\bm{p}_j + d_{ij} < -\frac{r_e}{2} \quad \forall \bm{p}_j \in {\mathcal{P}_j}. 
    \label{eq: collision}
    \end{aligned}
\end{equation}
The aforementioned equation guarantees that the distance between two agents over the period $\delta t$ exceeds the influence distance. Similarly, with the assumption that the $k^{th}$ static obstacle can be represented as a convex polygon with the vertex set $Q_k$, the collision-free constraints for this obstacle are expressed as
\begin{equation}
    \begin{aligned}
    &\bm{n}_{ik}^{\top}\bm{p}_i + d_{ik} > {r_a} \quad \forall \bm{p}_i \in {\mathcal{P}_i},  \\
    &\bm{n}_{ik}^{\top}\bm{q}_k + d_{ik} < -{r_a} \quad \forall \bm{q}_k \in {\mathcal{Q}_k}, 
    \label{eq: collision_static}
    \end{aligned}
\end{equation}
where $r_a$ denotes the radius of the agent. For circular obstacles, the constraints are simplified to distance constraints as
\begin{equation}
\|\bm{p}_{i} - \bm{x}_c\|_2 \leq r_c + {r_a}\quad \forall \bm{p}_i \in {\mathcal{P}_i},   
\label{eq: collision_static_2}
\end{equation}
where $\bm{x}_c$ and $r_c$ represent the obstacle position and radius, respectively.

\textit{2) Optimal Problem}: As the environment is fully observable, we adopt a centralized approach to optimize the trajectories of all agents simultaneously. %, which offers the advantage of optimality compared to distributed methods \cite{}. 
According to the dynamics and collision-free constraints, we formulate the trajectory generation as a nonlinear optimization problem: %The position of robots can be obtained by
% \begin{equation}
%     \bm{x}_o = \bm{G} \bm{x}_s =     \begin{bmatrix}
%        1 & 0 &  -d \cos{\theta} \\
%        0 & 1 &  -d \sin{\theta}
%     \end{bmatrix} \bm{x}_s,
%     \label{eq: simplified_dynamic}
% \end{equation}
% and the optimization problem is formulated as
\begin{equation}
\begin{aligned}
     \min\limits_{\mathcal{U}=\{\bm{u}_0, \cdots , \bm{u}_{L-1}\}} &\mathcal{J} = \sum_{i=0}^{N} \sum_{k=0}^{L_{i}}  \bm \|\bm{x}_{o,k+1} - \bm {x}_{o,k}\| \\ & +  \sum_{i=0}^{N} \bm \|\bm{x}_{o,L_i} - \bm {x}_{o,\text{targ}}\|_{\bm{W}} \\ 
    s.t. \quad &\bm{x}_{s,k+1} = \bm{f}_s^{\text{RK4}} (\bm{{x}}_{s,k}, \bm{{u}}_{s,k}, \delta t), \\
    &\bm{x}_{r,k} = \bm{G} \bm{x}_{s,k}, \bm{x}_0 = \bm{x}_{\text{init}},\quad k = 0, \cdots ,L_i \\
    %\bm{n}_{ij}^{\top}\bm{p}_i + d_{ij} > &\frac{r_c}{2} \quad \forall \bm{p}_i \in {\mathcal{P}_i},  \\
    %\bm{n}_{ij}^{\top}\bm{p}_j + d_{ij} < &-\frac{r_c}{2} \quad \forall \bm{p}_j \in {\mathcal{P}_j}.
    &\text{Collision}  \text{-free Constraints} (\ref{eq: collision}), (\ref{eq: collision_static})/ (\ref{eq: collision_static_2}).
\end{aligned}
\label{eq: optimal}
\end{equation}
The final position is added to the objective function as a penalty loss weighted with parameter $\bm{W}$ instead of a hard constraint to increase the degree of freedom of the problem. The trajectory generated by the global planner is employed as an initial feasible solution for the optimization problem, which is then solved using IPOPT \cite{wachter2006interior}, thereby significantly improving the solution efficiency.

\section{Results}
% \subsection{System Setup}
% The projection accuracy of the optoelectronic tweezers is crucial for the successful completion of manipulation tasks. 
We develop an OET manipulation platform that realizes closed-loop control with high projection precision and a wide field of view\footnote{Due to space limitations, the images of OET platform are presented in the supplementary video.}. The platform consists of an optical system, a motion stage, a microfluidic chip, and an AC generator. The projection system is formed by a digital micromirror device (DMD) projector (TI DLP6500) coupled with a projection lens, and it precisely projects light patterns generated on the PC side onto the chip. The observation pathway consists of a 20X objective lens (Olympus UPLFLN20X), a tube lens (SWTLU-C), and a 4K CCD camera (LBAS U3120-23C), and it is utilized for capturing microscope images with a wide field of view and transmitting them to the PC. 

A microfluidic space with a height of 100 µm is formed between the bottom and top chips, and the chips are prepared via plasma-enhanced chemical vapor deposition for manipulation tasks. Metallo-dielectric Janus particles are employed as robots and target objects driven by the OETs. We manually select the robot and the target object first and then utilize channel and spatial relatibility tracking in OpenCV \cite{bradski2000opencv} to maintain stable tracking under varying illumination. Both simulation and experimental algorithms are written in Python and run on an Intel i7 CPU (2.30GHz). 
\begin{figure}[tb]
    \vspace{-7pt}
    \centering
    \includegraphics[width=7cm]{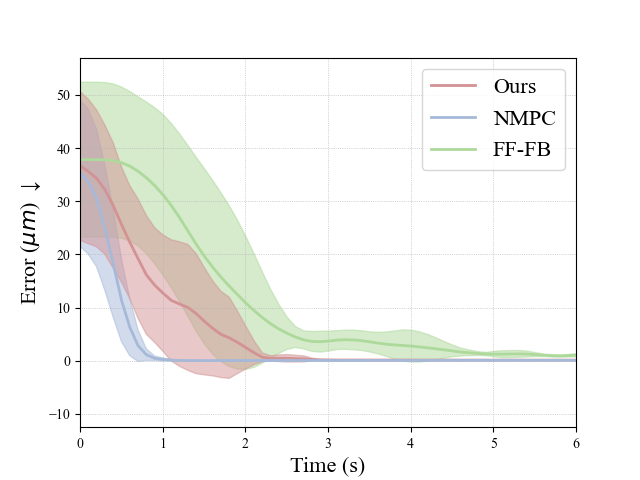}
    \vspace{-8pt}
    \caption{Variation in tracking error over time for different controllers in simulations. The mean values (solid lines) and standard deviation ranges (shaded areas) are shown.}
    \vspace{-12pt}
    \label{fig:error_sim}
\end{figure}
\vspace{-10pt}
\subsection{Trajectory Tracking}
We initially evaluate the performance of the trajectory tracking controller described in section III.B for the single non-contact trajectory tracking task in the numerical simulation. Two baseline controllers are realized: 1) the FF--FB controller defined in (\ref{eq: p_controller}) and 2) the nonlinear model predictive controller (NMPC) constrained by nonlinear dynamics in (\ref{eq: whole_nonlinear}), solved using CasADi \cite{Andersson2019}. The controller hyperparameters are set as $\bm{K}_p = \text{diag}(2,2,2)$, $H=5$, $\bm{Q}=\text{diag}(10,10,10)$, and $\bm{R}=\text{diag}(0.1,0.1,0.1)$. The target trajectory is defined as a circle with a radius of $40 
 \mu m$. Each controller is tested ten times, and the trajectory tracking error over time is recorded as shown in Fig. \ref{fig:error_sim}. The FF--FB controller features a low error convergence rate and a steady-state error owing to its neglect of nonlinearity. NMPC can ensure optimality within a finite horizon, but its average solution time of $0.171s$ makes it unsuitable for online deployment. Our method balances controller performance and solution time, reducing the average single-step solution time to $0.018s$, which helps increase the closed-loop control frequency and thus overcome the effects of self-propulsion.

In real-world experiments, we first test the tracking performance of a Cartesian curve trajectory, defined as
\begin{equation}
    \bm{x}_o^{\text{ref}} = 
    \begin{bmatrix}
    \frac{a \cos(\theta)}{1 + \sin(\theta) \sin(\theta)}  \\[10pt]
    \frac{b \sin(\theta) \cos(\theta)}{1 + \sin(\theta) \sin(\theta)}
    \end{bmatrix} \quad \hspace{-0.3cm}\theta \in \left[0, 2\pi\right], a = 160 \mu m, b = 160 \mu m.
    \label{eq: ref_traj}
\end{equation}
\begin{figure}[tb]
    \vspace{-5pt}
    \centering
    \includegraphics[width=8cm]{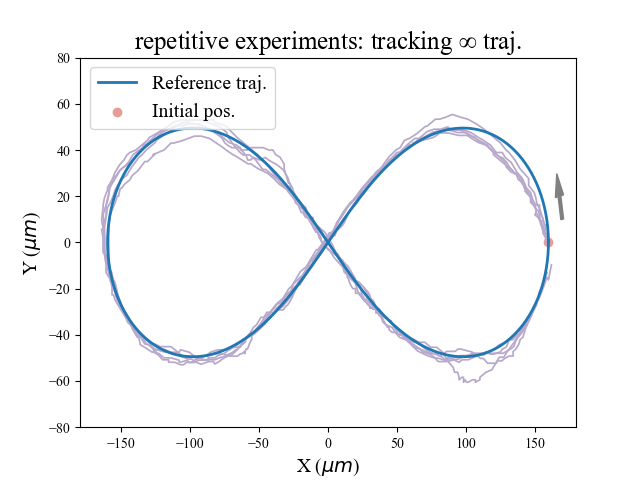}
    \vspace{-5pt}
    \caption{Five repeated experiments of curve tracking. The actual trajectory is depicted in light purple.}
    \vspace{-5pt}
    \label{fig:traj}
\end{figure}
\begin{figure}[tb]
    \vspace{-5pt}
    \centering
    \includegraphics[width=8cm]{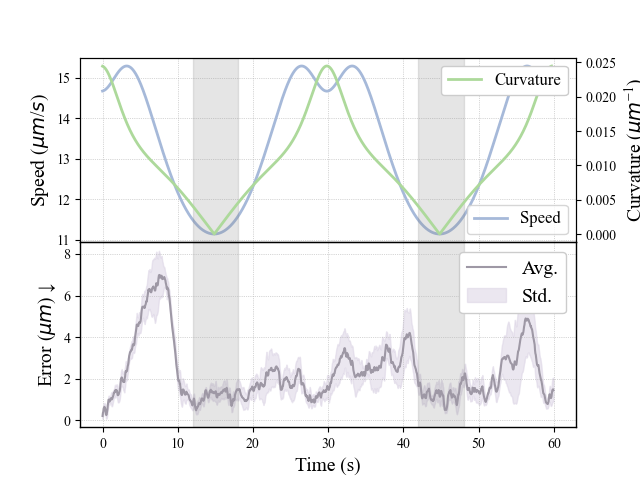}
    \vspace{-8pt}
    \caption{Relationships between tracking error and target speed and curvature. The gray areas indicate periods with smaller average errors.}
    \vspace{-8pt}
    \label{fig:error_exp}
\end{figure}
\begin{figure}[htb]
    \vspace{-5pt}
    \centering
    \includegraphics[width=8.5cm]{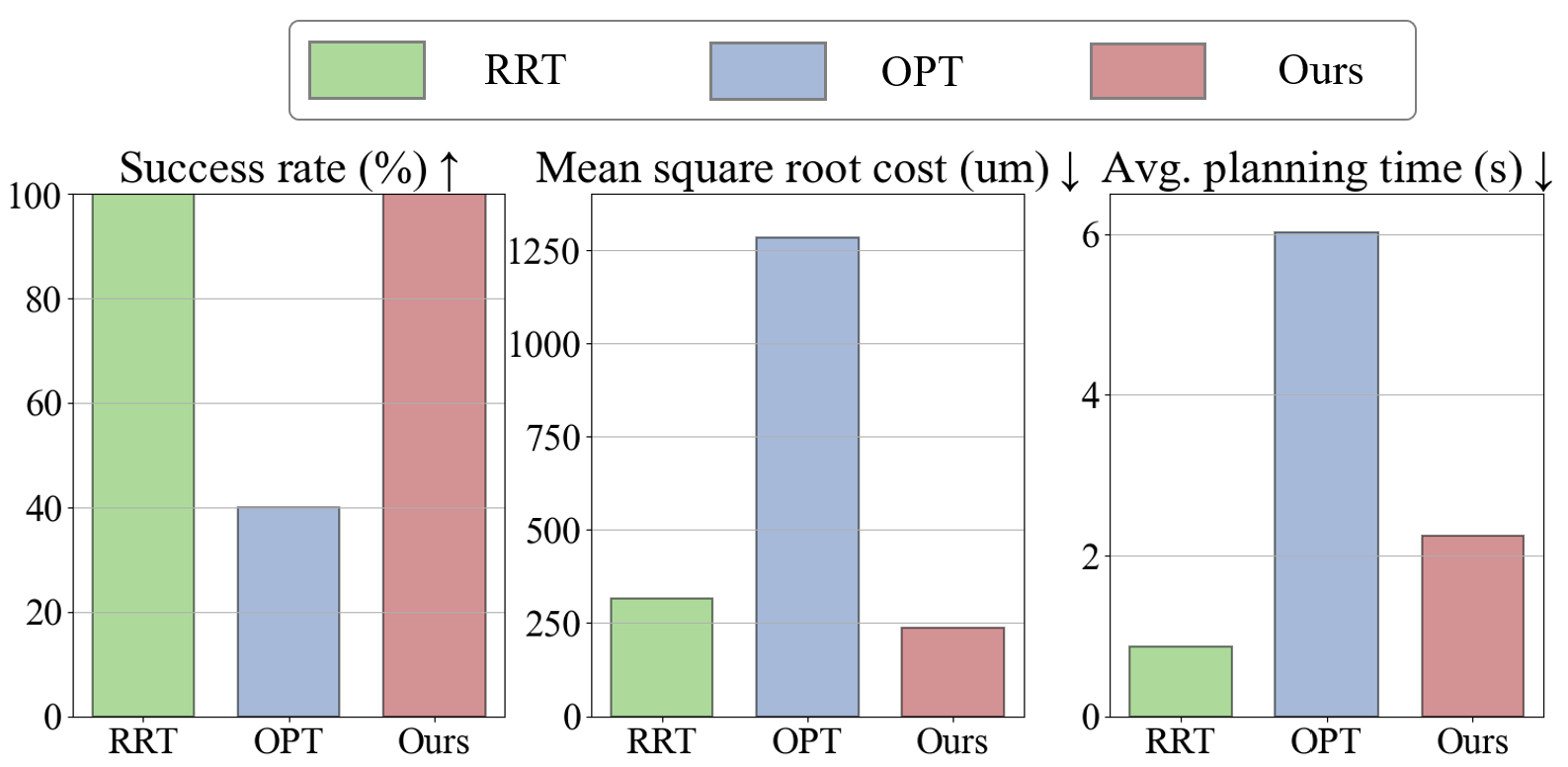}
    \vspace{-8pt}
    \caption{Comparison between the RRT-based planning scheme, the optimization-based planning scheme, and our planning scheme in simulation. The costs are computed using the objective function in (\ref{eq: optimal}).}
    \label{fig:table}
\end{figure}

As illustrated in Fig. \ref{fig:traj}, in five repetitive experiments, the target object robustly tracks the trajectory with a maximum error of $11.47 \mu m$  while maintaining the non-contact state with the robot. Fig. \ref{fig:error_exp} demonstrates the relationship between the tracking error, target trajectory curvature, and speed. The valleys in the tracking error correspond to periods of minimal speed and curvature, while the peaks correspond to higher speed and curvature periods. We quantitatively measure the average tracking error for circular trajectories under different radii and speeds (TABLE \ref{tab: radius}). The tracking error increases with speed, and the coefficient $\bm{K}_p$ accordingly increases to achieve better tracking performance.

\begin{figure*}[bth]
    \vspace{-4pt}
    \centering
    \includegraphics[width=15.5cm]{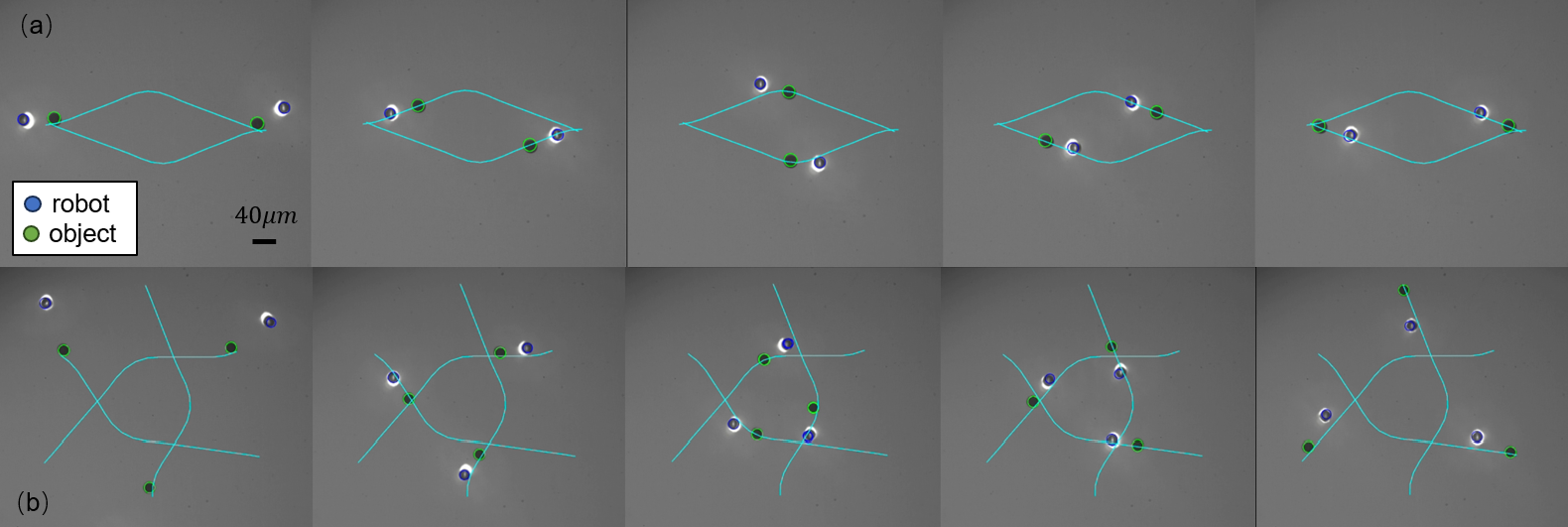}
    \vspace{-8pt}
    \caption{Snapshots of obstacle avoidance between multiple robots and objects: (a) two robot--object pairs; (b) three robot--object pairs. The robots and target objects are outlined in blue and green circles. The blue line represents the target trajectory planned by our method. The first and last columns of images show the initial and target states, respectively. }
    \label{fig:agent}
    \vspace{-14pt}
\end{figure*}

\begin{table}[ht]
\vspace{-8pt}
\caption{Mean Tracking Error of Circular Trajectory}
 \vspace{-8pt}
    \begin{center} 
    \begin{threeparttable}
    \centering
        \begin{tabular}{ccccc}
        % \hline
        \toprule[0.8pt]
        \multicolumn{2}{c}{\multirow{2}{*}{Mean error ($\mu m$) $\downarrow$}}   & \multicolumn{3}{c}{Radius ($\mu m$)}   \\ 
         &  & {100}       & {160}      & {200}            \\ 
        % \hline
        \midrule[0.5pt]
        \multirow{3}{*}{Speed ($\mu m /s$)}      
        & 10  & 2.010  & 1.488   & 2.028      \\ 
        & 16  & 5.534  & 2.756   & 2.098      \\
        & 20  & 9.322  & 10.38   & 4.146      \\  
       \bottomrule[0.8pt] 
       \end{tabular}
    \end{threeparttable}
    \end{center} 
    \label{tab: radius}
 \vspace{-12pt}
\end{table}

% \begin{table}[ht]
% \caption{Average Single-step Solution Time}

%     \begin{center} 
%     \begin{threeparttable}
%     \centering
%         \begin{tabular}{ccccc}
%         % \hline
%         \toprule[0.8pt]
   
%         {\multirow{2}{*}{Mean Time (ms) $\downarrow$}}   & \multicolumn{3}{c}{Method}   \\ 
%          & {Feedback}       &  {Ours}    &  {NMPC}         \\ 
%         % \hline
%         \midrule[0.5pt]
%         % \multirow{2}{*}{Speed}      
%          line  & 0.39  & 18   & 171     \\ 
%          circle  & 0.39  & 23   & 162      \\
%        \bottomrule[0.8pt] 
%        \end{tabular}
%     \end{threeparttable}
%     \end{center} 
%     \label{tab: radius}
% \end{table}

\subsection{Multi-agent Navigation}
We validate the non-contact multi-agent planning method proposed in Section IV. The parameters are selected as $r_e = 6r_a$, $\bm{W} = \text{diag}(5,5)$, and $\delta t = 2s$. In the scenario without static obstacles, we compare our method with the distributed multi-agent obstacle avoidance method, RVO. Under each number of agent pairs, 20 experiments are conducted with randomized initial and target positions. As depicted in Table \ref{tab: rvo}, our planner achieves shorter paths through the simultaneous optimization of the trajectories of all agents across all time steps. Moreover, the advantage becomes more pronounced as the number of agents increases. 

Ablation experiments are further executed to compare our coarse-to-fine framework with two baseline methods: 1) RRT, RRT-based global planning described in Section IV.A, and 2) OPT, in which the optimization problem in Section IV.B is directly solved without an initial solution. Ten simulations are performed with randomized initial positions in a scenario involving three pairs of agents and three obstacles. As shown in Fig. \ref{fig:table}, owing to the highly nonlinear nature of the problem, the direct optimization method is prone to converge to the point of local infeasibility, leading to planning failures and incurring high computational costs. In contrast, our global planner first provides a near-feasible initial trajectory, which is further refined through local optimization to shorten the trajectory length. The total time cost for both stages is approximately 2.2 s, demonstrating high planning efficiency.

\begin{figure}[htb]
    \vspace{-10pt}
    \centering
    \includegraphics[width=8cm]{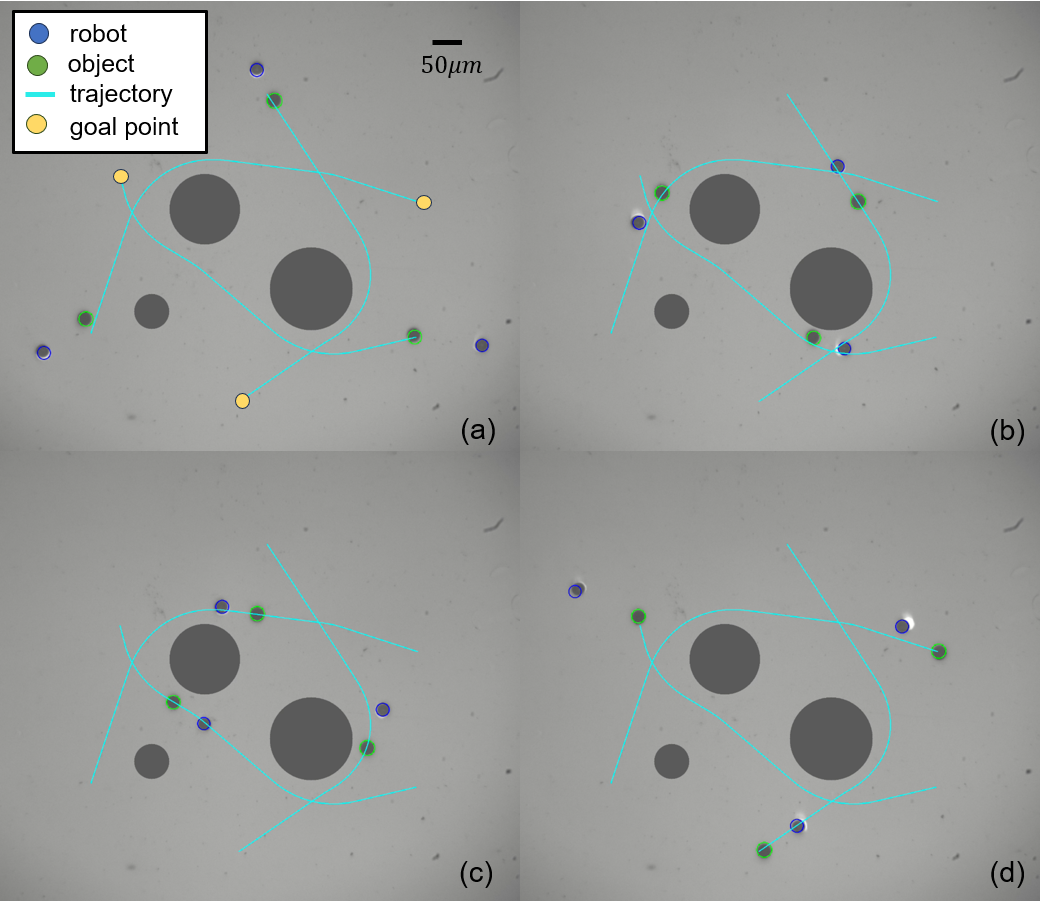}
    \vspace{-8pt}
    \caption{Snapshots of multi-agent non-contact navigation in the presence of static obstacles. The gray circles represent three virtual circular obstacles. The blue line represents the target trajectory planned by our method. (a) 0 s (b) 21 s (c) 36 s (d) 57 s }
    \label{fig:agent_2}
    \vspace{-8pt}
\end{figure}

Finally, we demonstrate the multi-agent navigation performance in the real-world environment. As shown in Fig. \ref{fig:agent}, in scenarios where the straight paths to the targets intersect, two and three robots can transport objects to their target positions without contact, effectively showcasing inter-robot and object avoidance. Fig. \ref{fig:agent_2} illustrates obstacle avoidance in the presence of static obstacles. The proposed framework can plan smooth trajectories that navigate around virtual obstacles and avoid collisions between agents. The controller robustly tracks these trajectories, ensuring the successful completion of the multi-agent navigation task.

\begin{table}[ht]
\caption{Average Trajectory Length of Multi-agent Planning}
    \vspace{-8pt}
    \begin{center} 
    \begin{threeparttable}
    \centering
        \begin{tabular}{ccccc}
        % \hline
        \toprule[0.8pt]
        \multicolumn{2}{c}{\multirow{2}{*}{Length ($mm$)$ \downarrow$}}   & \multicolumn{3}{c}{Number of robot--object pairs}   \\ 
         &     & {3}      & {4}       &{5}    \\ 
        % \hline
        \midrule[0.5pt]
        \multirow{2}{*}{Method}      
        & RVO  & $1.827 \pm 0.458 $ & $2.961\pm0.585$ & $3.308 \pm 0.638$ \\ 
        & Ours  & $ 1.818 \pm 0.560$  & $2.699 \pm 0.494$   & $2.661 \pm 0.561$   \\
       \bottomrule[0.8pt] 
       \end{tabular}
    \end{threeparttable}
    \end{center} 
    \label{tab: rvo}
    \vspace{-5pt}
\end{table}

\section{Conclusions}
This paper proposes a novel non-contact dexterous micromanipulation method with multiple optoelectronic robots and designs a comprehensive planning and control framework. The universal microrobots, captured by light spots, utilize dielectrophoretic repulsive forces to push target objects in a non-contact manner, offering dexterous and non-destructive characteristics. A linear MPC controller is constructed using two motion models to balance the trade-off between the long horizon and control frequency and achieve online trajectory tracking for underactuated systems. Furthermore, a coarse-to-fine multi-agent planning framework is proposed to navigate multiple robot--object pairs in cluttered scenarios. Simulations and experiments demonstrate that our framework can efficiently plan and track collision-free trajectories, completing dexterous manipulation tasks without contact. Future work will focus on achieving online planning for dynamic obstacles and applying the planning to real-world cell manipulation.

{\small
\bibliographystyle{IEEEtran}
\bibliography{ref}

\begin{thebibliography}{10}
\providecommand{\url}[1]{#1}
\csname url@samestyle\endcsname
\providecommand{\newblock}{\relax}
\providecommand{\bibinfo}[2]{#2}
\providecommand{\BIBentrySTDinterwordspacing}{\spaceskip=0pt\relax}
\providecommand{\BIBentryALTinterwordstretchfactor}{4}
\providecommand{\BIBentryALTinterwordspacing}{\spaceskip=\fontdimen2\font plus
\BIBentryALTinterwordstretchfactor\fontdimen3\font minus \fontdimen4\font\relax}
\providecommand{\BIBforeignlanguage}[2]{{%
\expandafter\ifx\csname l@#1\endcsname\relax
\typeout{** WARNING: IEEEtran.bst: No hyphenation pattern has been}%
\typeout{** loaded for the language `#1'. Using the pattern for}%
\typeout{** the default language instead.}%
\else
\language=\csname l@#1\endcsname
\fi
#2}}
\providecommand{\BIBdecl}{\relax}
\BIBdecl

\bibitem{xu2018micromachines}
Q.~Xu, \emph{Micromachines for biological micromanipulation}.\hskip 1em plus 0.5em minus 0.4em\relax Springer, 2018.

\bibitem{ahmad2021mobile}
B.~Ahmad, M.~Gauthier, G.~J. Laurent, and A.~Bolopion, ``Mobile microrobots for in vitro biomedical applications: A survey,'' \emph{IEEE Transactions on Robotics}, vol.~38, no.~1, pp. 646--663, 2021.

\bibitem{miao2023microfluidics}
S.~Miao, J.~Xu, Z.~Jiang, J.~Luo, X.~Sun, X.~Jiang, H.~Wei, and Y.-H. Liu, ``Microfluidics-enabled robotic system for embryo vitrification with real-time observation: design, method, and evaluation,'' \emph{IEEE/ASME Transactions on Mechatronics}, vol.~29, no.~1, pp. 179--189, 2023.

\bibitem{dai2022robotic}
C.~Dai, G.~Shan, H.~Liu, C.~Ru, and Y.~Sun, ``Robotic manipulation of sperm as a deformable linear object,'' \emph{IEEE Transactions on Robotics}, vol.~38, no.~5, pp. 2799--2811, 2022.

\bibitem{sun2022robotic}
D.~Sun, \emph{Robotic Cell Manipulation}.\hskip 1em plus 0.5em minus 0.4em\relax Academic Press, 2022.

\bibitem{zhang2022optoelectronic}
S.~Zhang, B.~Xu, M.~Elsayed, F.~Nan, W.~Liang, J.~K. Valley, L.~Liu, Q.~Huang, M.~C. Wu, and A.~R. Wheeler, ``Optoelectronic tweezers: a versatile toolbox for nano-/micro-manipulation,'' \emph{Chemical Society Reviews}, vol.~51, no.~22, pp. 9203--9242, 2022.

\bibitem{optoelectronic_microrobot}
S.~Zhang, E.~Scott, J.~Singh, Y.~Chen, Y.~Zhang, M.~Elsayed, M.~Chamberlain, N.~Shakiba, K.~Adams, S.~Yu, C.~Morshead, P.~Zandstra, and A.~Wheeler, ``The optoelectronic microrobot: A versatile toolbox for micromanipulation,'' \emph{Proceedings of the National Academy of Sciences}, vol. 116, p. 201903406, 07 2019.

\bibitem{Massively_parallel}
P.-Y. Chiou, A.~Ohta, and M.~Wu, ``Massively parallel manipulation of single cells and microparticles using optical images,'' \emph{Nature}, vol. 436, pp. 370--2, 08 2005.

\bibitem{Trap_profiles}
S.~Neale, A.~Ohta, H.-Y. Hsu, J.~Valley, A.~Jamshidi, and M.~Wu, ``Trap profiles of projector based optoelectronic tweezers (oet) with hela cells,'' \emph{Optics express}, vol.~17, pp. 5232--9, 04 2009.

\bibitem{single_DNA}
L.~Yen-Hengg, C.-M. Chang, and G.-B. Lee, ``Manipulation of single dna molecules by using optically projected images,'' \emph{Optics express}, vol.~17, pp. 15\,318--29, 09 2009.

\bibitem{mi11010078}
\BIBentryALTinterwordspacing
W.~Liang, L.~Liu, J.~Wang, X.~Yang, Y.~Wang, W.~J. Li, and W.~Yang, ``A review on optoelectrokinetics-based manipulation and fabrication of micro/nanomaterials,'' \emph{Micromachines}, vol.~11, no.~1, 2020. [Online]. Available: \url{https://www.mdpi.com/2072-666X/11/1/78}
\BIBentrySTDinterwordspacing

\bibitem{Optoelectrokinetics}
W.~Liang, L.~Liu, J.~Wang, X.~Yang, Y.~Wang, W.~Li, and W.~Yang, ``A review on optoelectrokinetics-based manipulation and fabrication of micro/nanomaterials,'' \emph{Micromachines}, vol.~11, p.~78, 01 2020.

\bibitem{cenev2021ferrofluidic}
Z.~Cenev, P.~D. Harischandra, S.~Nurmi, M.~Latikka, V.~Hynninen, R.~H. Ras, J.~V. Timonen, and Q.~Zhou, ``Ferrofluidic manipulator: Automatic manipulation of nonmagnetic microparticles at the air--ferrofluid interface,'' \emph{IEEE/ASME Transactions on Mechatronics}, vol.~26, no.~4, pp. 1932--1940, 2021.

\bibitem{kim2021measurements}
S.~Kim, S.~Moon, S.~Rho, and S.~Yoon, ``Measurements of acoustic radiation force of ultrahigh frequency ultrasonic transducers using model-based approach,'' \emph{Applied physics letters}, vol. 118, no.~18, 2021.

\bibitem{icsitman2021non}
O.~I{\c{s}}{\i}tman, H.~Bettahar, and Q.~Zhou, ``Non-contact cooperative manipulation of magnetic microparticles using two robotic electromagnetic needles,'' \emph{IEEE Robotics and Automation Letters}, vol.~7, no.~2, pp. 1605--1611, 2021.

\bibitem{lim2013ultrahigh}
S.~W. Lim and A.~R. Abate, ``Ultrahigh-throughput sorting of microfluidic drops with flow cytometry,'' \emph{Lab on a Chip}, vol.~13, no.~23, pp. 4563--4572, 2013.

\bibitem{10610098}
Y.~Jia, S.~Miao, J.~Zhou, N.~Jiao, L.~Liu, and X.~Li, ``Efficient model learning and adaptive tracking control of magnetic micro-robots for non-contact manipulation,'' in \emph{2024 IEEE International Conference on Robotics and Automation (ICRA)}, 2024, pp. 4534--4540.

\bibitem{fan2018automated}
X.~Fan, M.~Sun, Z.~Lin, J.~Song, Q.~He, L.~Sun, and H.~Xie, ``Automated noncontact micromanipulation using magnetic swimming microrobots,'' \emph{IEEE Transactions on Nanotechnology}, vol.~17, no.~4, pp. 666--669, 2018.

\bibitem{huang2017path}
L.~Huang, L.~Rogowski, M.~J. Kim, and A.~T. Becker, ``Path planning and aggregation for a microrobot swarm in vascular networks using a global input,'' in \emph{2017 IEEE/RSJ International Conference on Intelligent Robots and Systems (IROS)}.\hskip 1em plus 0.5em minus 0.4em\relax IEEE, 2017, pp. 414--420.

\bibitem{wang2021micromanipulation}
Q.~Wang, L.~Yang, and L.~Zhang, ``Micromanipulation using reconfigurable self-assembled magnetic droplets with needle guidance,'' \emph{IEEE Transactions on Automation Science and Engineering}, vol.~19, no.~2, pp. 759--771, 2021.

\bibitem{liu20203}
J.~Liu, X.~Wu, C.~Huang, L.~Manamanchaiyaporn, W.~Shang, X.~Yan, and T.~Xu, ``3-d autonomous manipulation system of helical microswimmers with online compensation update,'' \emph{IEEE Transactions on Automation Science and Engineering}, vol.~18, no.~3, pp. 1380--1391, 2020.

\bibitem{202100279}
J.~Jiang, Z.~Yang, A.~Ferreira, and L.~Zhang, ``Control and autonomy of microrobots: Recent progress and perspective,'' \emph{Advanced Intelligent Systems}, vol.~4, no.~5, p. 2100279, 2022.

\bibitem{yang2019automated}
L.~Yang, Y.~Zhang, Q.~Wang, K.-F. Chan, and L.~Zhang, ``Automated control of magnetic spore-based microrobot using fluorescence imaging for targeted delivery with cellular resolution,'' \emph{IEEE Transactions on Automation Science and Engineering}, vol.~17, no.~1, pp. 490--501, 2019.

\bibitem{lee2021real}
J.~Lee, X.~Zhang, C.~H. Park, and M.~J. Kim, ``Real-time teleoperation of magnetic force-driven microrobots with 3d haptic force feedback for micro-navigation and micro-transportation,'' \emph{IEEE Robotics and Automation Letters}, vol.~6, no.~2, pp. 1769--1776, 2021.

\bibitem{li2018development}
X.~Li, S.~Chen, C.~Liu, S.~H. Cheng, Y.~Wang, and D.~Sun, ``Development of a collision-avoidance vector based control algorithm for automated in-vivo transportation of biological cells,'' \emph{Automatica}, vol.~90, pp. 147--156, 2018.

\bibitem{3263773}
Y.~Liu, H.~Chen, Q.~Zou, X.~Du, Y.~Wang, and J.~Yu, ``Automatic navigation of microswarms for dynamic obstacle avoidance,'' \emph{IEEE Transactions on Robotics}, vol.~PP, pp. 1--16, 08 2023.

\bibitem{9636475}
C.~Bendkowski, L.~Mennillo, T.~Xu, M.~Elsayed, F.~Stojic, H.~Edwards, S.~Zhang, C.~Morshead, V.~Pawar, A.~R. Wheeler, D.~Stoyanov, and M.~Shaw, ``Autonomous object harvesting using synchronized optoelectronic microrobots,'' in \emph{2021 IEEE/RSJ International Conference on Intelligent Robots and Systems (IROS)}, 2021, pp. 7498--7504.

\bibitem{9841614}
L.~Mennillo, C.~Bendkowski, M.~Elsayed, H.~Edwards, S.~Zhang, V.~Pawar, A.~R. Wheeler, D.~Stoyanov, and M.~Shaw, ``Adaptive autonomous navigation of multiple optoelectronic microrobots in dynamic environments,'' \emph{IEEE Robotics and Automation Letters}, vol.~7, no.~4, pp. 11\,102--11\,109, 2022.

\bibitem{Electrostatic_Force}
H.~TANG, L.~Gan, and W.~AN, ``Electrostatic force between two dielectric particles in electrorheological fluids: beyond spherical particles,'' \emph{MATEC Web of Conferences}, vol. 187, p. 04001, 08 2018.

\bibitem{hogan2020reactive}
F.~R. Hogan and A.~Rodriguez, ``Reactive planar non-prehensile manipulation with hybrid model predictive control,'' \emph{The International Journal of Robotics Research}, vol.~39, no.~7, pp. 755--773, 2020.

\bibitem{10341476}
Y.~Jiang, Y.~Jia, and X.~Li, ``Contact-aware non-prehensile manipulation for object retrieval in cluttered environments,'' in \emph{2023 IEEE/RSJ International Conference on Intelligent Robots and Systems (IROS)}, 2023, pp. 10\,604--10\,611.

\bibitem{10.5555/3165183}
K.~M. Lynch and F.~C. Park, \emph{Modern Robotics: Mechanics, Planning, and Control}, 1st~ed.\hskip 1em plus 0.5em minus 0.4em\relax USA: Cambridge University Press, 2017.

\bibitem{10611608}
A.~Wang, C.~Gan, H.~Han, H.~Xiong, J.~Zhao, C.~Wang, and L.~Feng, ``Dynamic adaptive imaging system on optoelectronic tweezers platform,'' in \emph{2024 IEEE International Conference on Robotics and Automation (ICRA)}, 2024, pp. 15\,622--15\,627.

\bibitem{tordesillas2019faster}
J.~Tordesillas, B.~T. Lopez, and J.~P. How, ``Faster: Fast and safe trajectory planner for flights in unknown environments,'' in \emph{2019 IEEE/RSJ international conference on intelligent robots and systems (IROS)}.\hskip 1em plus 0.5em minus 0.4em\relax IEEE, 2019, pp. 1934--1940.

\bibitem{zhou2019robust}
B.~Zhou, F.~Gao, L.~Wang, C.~Liu, and S.~Shen, ``Robust and efficient quadrotor trajectory generation for fast autonomous flight,'' \emph{IEEE Robotics and Automation Letters}, vol.~4, no.~4, pp. 3529--3536, 2019.

\bibitem{wachter2006interior}
A.~W{\"a}chter and L.~T. Biegler, ``On the implementation of an interior-point filter line-search algorithm for large-scale nonlinear programming,'' \emph{Mathematical Programming}, vol. 106, pp. 25--57, 2006.

\bibitem{bradski2000opencv}
G.~Bradski, ``The opencv library.'' \emph{Dr. Dobb's Journal: Software Tools for the Professional Programmer}, vol.~25, no.~11, pp. 120--123, 2000.

\bibitem{Andersson2019}
J.~A.~E. Andersson, J.~Gillis, G.~Horn, J.~B. Rawlings, and M.~Diehl, ``{CasADi} -- {A} software framework for nonlinear optimization and optimal control,'' \emph{Mathematical Programming Computation}, vol.~11, no.~1, pp. 1--36, 2019.

\end{thebibliography}
}
\end{document}